\documentclass{ieeeaccess}
\usepackage[numbers,sort]{natbib}
\usepackage{amsmath,amssymb,amsfonts}
\usepackage{algorithmic}
\usepackage{graphicx}
\usepackage{textcomp}
\usepackage{multirow} 
\usepackage{booktabs} 

\def\BibTeX{{\rm B\kern-.05em{\sc i\kern-.025em b}\kern-.08em
    T\kern-.1667em\lower.7ex\hbox{E}\kern-.125emX}}
\begin{document}
\history{Date of publication xxxx 00, 2023, date of current version xxxx 00, 2023.}
\doi{10.1109/ACCESS.2023.0092316}

\title{Value Iteration Networks with Gated Summarization Module}
\author{\uppercase{Jinyu Cai}\authorrefmark{1},
\uppercase{Jialong Li}\authorrefmark{1},\IEEEmembership{Student Member, IEEE},
\uppercase{Mingyue Zhang}\authorrefmark{2},\IEEEmembership{Member, IEEE},
and 
\uppercase{Kenji Tei}\authorrefmark{1},\IEEEmembership{Member, IEEE}}

\address[1]{Faculty of Science and Engineering, Waseda University, 3-4-1 Okubo, Shinjuku, Tokyo 169-8555, Japan}
\address[2]{College of Computer and Information science, Southwest University, No.2 Tiansheng Road, Beibei District, Chongqing, China, 400715}

\tfootnote{This work was partially supported by JSPS KAKENHI.}

\markboth
{Cai \headeretal: Value Iteration Networks with Gated Summarization Module}
{Cai \headeretal: Value Iteration Networks with Gated Summarization Module}

\corresp{Corresponding author: Jinyu Cai (e-mail: bluelink@toki.waseda.jp).}

\begin{abstract}
In this paper, we address the challenges faced by Value Iteration Networks (VIN) in handling larger input maps and mitigating the impact of accumulated errors caused by increased iterations. We propose a novel approach, Value Iteration Networks with Gated Summarization Module (GS-VIN), which incorporates two main improvements: (1) employing an Adaptive Iteration Strategy in the Value Iteration module to reduce the number of iterations, and (2) introducing a Gated Summarization module to summarize the iterative process. The adaptive iteration strategy uses larger convolution kernels with fewer iteration times, reducing network depth and increasing training stability while maintaining the accuracy of the planning process. The gated summarization module enables the network to emphasize the entire planning process, rather than solely relying on the final global planning outcome, by temporally and spatially resampling the entire planning process within the VI module. We conduct experiments on 2D grid world path-finding problems and the Atari Mr. Pac-man environment, demonstrating that GS-VIN outperforms the baseline in terms of single-step accuracy, planning success rate, and overall performance across different map sizes. Additionally, we provide an analysis of the relationship between input size, kernel size, and the number of iterations in VI-based models, which is applicable to a majority of VI-based models and offers valuable insights for researchers and industrial deployment.
\end{abstract}


\titlepgskip=-15pt

\maketitle

\section{Introduction}
In path-finding problems, intelligent agents need to find the optimal path from a starting point to an endpoint in a given environment. 
These problems have significant implications in various practical applications such as autonomous driving, robot navigation, and gaming AI. 
To address these issues, researchers have proposed many different algorithms and models, such as the Dijkstra's algorithm\cite{dijkstra}, A* algorithm\cite{Astar}, etc. 
However, these algorithms usually require global environmental information and have low computational efficiency in complex environments. 
To address the computational efficiency issues of traditional path-finding algorithms in complex environments, Tamar et al. proposed Value Iteration Networks (VIN) in 2016\cite{VIN}. 
VIN is an end-to-end trainable neural network that combines the concepts of Value Iteration (VI)\cite{Bellman1957} and Convolutional Neural Networks (CNN)\cite{LeCun1998}. 
The core idea of VIN is to embed the Value Iteration algorithm as a planning module (also called the value iteration module) into the neural network architecture to perform active planning. 
This design allows VIN to directly learn the optimal policy from the input's raw information through active global planning without explicit state space modeling or pre-defined feature representations. 
According to literature \cite{VIN}, it has been demonstrated that VIN exhibits superior performance and generalization capabilities when compared to conventional reactive neural networks.

Although VIN has achieved some success, it still faces many challenges. 
Recent research based on VIN has addressed many of these issues from different perspectives, such as improving network overestimation\cite{DVIN}, enhancing VIN's generalization capabilities\cite{TVIN}, and enabling networks to handle larger map inputs through multi-sampling of the input\cite{AVIN}.
However, in these studies, to the best of our knowledge, researchers often overlook the utilization of convolutional layers within the VI module, primarily employing conventional $3 \times 3$ kernel sizes\cite{DVIN,TVIN,AVIN,UVIN,Nardelli2018ValuePN,Pflueger2018RoverIRLIR,sykora2020multi}.
This decision results in a prevalent issue among such networks: when confronted with larger inputs, the network can only achieve global planning through an increased number of iterations. Consequently, this necessitates a deeper network depth, which ultimately leads to the excessive accumulation of single-iteration computation errors and training instability stemming from the network's depth and the backpropagation algorithm\cite{rumelhart1986learning}, such as vanishing gradients and exploding gradients\cite{hochreiter2001gradient}.

Previous research \cite{VIRN} has provided preliminary investigations into the aforementioned problems. In this paper, we propose a more comprehensive approach, namely Value Iteration Networks with Gated Summarization Module (GS-VIN), building upon the previous research. The GS-VIN model includes two key enhancements: (1) adopting an adaptive iteration strategy in the value iteration (VI) module to reduce the number of iterations required, and (2) introducing a gated summarization(GS) module to robustly summarize the iterative process, thereby mitigating the impact of accumulated errors. These improvements aim to achieve more precise planning in larger and more complex environments.
The proposed improvements can be summarized as follows:
\begin{enumerate}
\item An adaptive iteration strategy in the VI module actively uses larger convolution kernels in conjunction with fewer iteration times. This approach not only reduces network depth, lowering the risk of gradient explosion and vanishing, and increasing training stability, but also ensures the accuracy of the planning process.
\item Recognizing the importance of considering the future prediction process when making decisions, rather than merely focusing on the ultimate future outcome, we propose a new gated summarization module. In larger inputs, the VI module requires more iterations, making the final iteration results more unreliable due to the accumulation of single-iteration errors. Therefore, our proposed module employs gating mechanisms to achieve a similar effect to attention mechanisms\cite{mnih2014recurrent}. With this module, the network can temporally and spatially sample the entire planning process within the VI module, rather than solely relying on the global planning output. This approach endows the model with the capacity to abstract and summarize the planning process, ultimately mitigating the impact of cumulative errors on the network's output.
\end{enumerate}

We conduct experiments on 2D grid world path-finding problems and the Atari Mr. Pac-man environment. 
For the 2D grid world path-finding problems, we demonstrate that GS-VIN outperforms the baseline in single-step accuracy and planning success rate across different map sizes. 
We also analyze the relationship between input size, convolution kernel size, and iteration times in the VI module of VI-based models based on 2D grid world task. 
In the Mr. Pac-man environment, the experiments show that GS-VIN maintains superior accuracy performance in larger and more complex environments compared to other competitors.

The main contributions of this paper are as follows: 
\begin{itemize}

    \item We propose the GS-VIN network, GS-VIN employs an adaptive iteration strategy in the value iteration module, and additionally incorporates a gated summarization module to mitigate error accumulation. These characteristics enable the network model to emphasize the entire planning process from local to global, rather than solely relying on the final global planning outcome, while simultaneously reducing the number of iterations in the VI module. Ultimately, this enhances the model's ability to summarize and improve the planning process. 

    \item We present a novel heuristic function to reveal the relationship among input size, convolution kernel size, and iteration number in adaptive iteration strategy, and conduct extensive experiments to investigate the impact of different combinations of convolution kernel size and iteration number on network performance. The experiment results indicate that the adaptive iteration strategy can be applied to VI-based network models, and the heuristic function we propose has an important reference value for VI-based network models. 

    
    \item We conduct experiments on 2D grid world path-finding problems and Atari's Mr. Pac-man environment, comparing GS-VIN with the method in \cite{VIRN} and other methods. We demonstrate that GS-VIN outperforms the baseline in terms of overall performance.

\end{itemize}
The contribution of this paper builds upon the direction proposed in prior work \cite{VIRN} and further investigates it. For a detailed comparison between this study and the work presented in \cite{VIRN}, please refer to section \ref{sec: Realted work}.
The structure of this paper is as follows: In section~\ref{sec: background}, we introduce some background knowledge related to this research. In section~\ref{sec: Realted work}, we review recent related work on VIN. In section~\ref{sec: Method}, we provide a detailed description of our method. In section~\ref{sec: Evaluation}, we present experiments and discuss the results. Finally, in section \ref{sec: Conclusion}, we summarize our work and mention future work.

\section{Background}\label{sec: background}

\subsection{Value Iteration}
Value iteration is a dynamic programming algorithm used to find the optimal policy in Markov Decision Processes (MDPs)\cite{Bellman1957}, by computing the optimal value function for each state and the optimal policy based on that value function. The algorithm initializes the value function of each state to 0 and iteratively updates it until convergence to the optimal solution. The basic idea of the algorithm is to start with a random policy and then use the Bellman optimality equation to iteratively update the value function of each state until it converges.

The value function of a state $s$, denoted by $V(s)$, represents the expected future reward starting from that state. The Bellman optimality equation states that the optimal value of a state can be calculated as the sum of the current reward and the maximum value of the next possible states, discounted by a factor $\gamma$ that represents the relative value of future rewards:

\begin{equation}\label{eq:vi} 
\begin{aligned}
  &V(s)=\max_{a} Q(s,a)\,\\
  &Q( s,a) = \sum _{s'}P( s'\vert s,a) [R(s,a,s') +\gamma V( s')]
\end{aligned}
\end{equation}

Here, $a$ denotes all possible actions that can be taken from state $s$, $P( s'\vert s, a)$ is the probability of transitioning to state $s'$ when taking action $a$ in state $s$, $R(s, a, s')$ represents the immediate reward received after transitioning from state $s$ to state $s'$ due to taking action $a$, and $\gamma$ is a discount factor that determines the weight of future rewards relative to current rewards.

In practice, we typically use iterative methods to solve the Bellman optimality equation. Starting from a zero-initialized value function $V_0$, the algorithm updates the value function for each state as follows:

\begin{equation}\label{eq:vi2} 
\begin{aligned}
  &V_{k+1}(s)=\max_{a} Q_k(s,a)\,\\
  &Q_{k}( s,a) = \sum _{s'}P( s'\vert s,a) [R(s,a,s') +\gamma V_{k}( s')]
\end{aligned}
\end{equation}

Here, $k$ denotes the iteration number, $V_k(s)$ represents the value function of state $s$ at the $k$-th iteration, and $Q_k(s, a)$ represents the action-value function at the $k$-th iteration for taking action $a$ in state $s$. It measures the expected long-term return of performing a specific action in a given state. The algorithm terminates when the value function converges to the optimal solution, and outputs the optimal value function and policy.


\subsection{Value Iteration Network}
Value Iteration Network \cite{VIN} is an end-to-end neural network used for solving optimal policies. 
It was proposed by Tamar et al. in 2016 and is based on the idea of using a convolutional neural network (CNN) to approximate the value iteration algorithm in the value iteration module (VI module), in order to improve its efficiency and accuracy.

In the standard value iteration algorithm, we need to iterate over every state until convergence. 
This algorithm is inefficient when dealing with large state spaces and requires significant computational resources and time. 
Therefore, VIN uses a CNN to approximate the value iteration process to improve efficiency. 
In the VI module, the value iteration algorithm is transformed into a continuous function that can be implemented by a CNN, and then a CNN is used to approximate this function iteratively, in order to obtain the optimal value function and policy. 
The value iteration algorithm in the VI module is expressed by the following equation:

\begin{equation}\label{eq:vin} 
\begin{aligned}
    V_{k+1}&=\max_{\alpha} Q^{\alpha}_{k}\\
    Q^{\alpha}_{k}&=W_{\alpha}^{R} * R+ W_{\alpha}^{V} * V_{k}\\
\end{aligned}
\end{equation}
where $W^{V}$,$W^{R}$ are the weights for the value function and reword function in the convolutional layer, $\alpha$ is the action index, $R$,$V$,$Q$ are estimated reward, estimated value function, estimated action-value function by neural networks, here ‘$*$’ denotes the convolution operator.
Same as the VI algorithm, the VI module also needs to iterate until $V$ reaches a steady state.

Value Iteration Network offers an innovative approach to designing end-to-end neural networks for solving optimal policies. 
Unlike previous reflexive neural networks, the iterative calculation in the value iteration module is an active planning process of the network to obtain the value function for input states, making VIN have superior generalization ability compared to other networks. 
In fact, in various practical applications, such as urban path planning\cite{Yang2018LearningUN}, planning of UAV swarms\cite{Li2021DynamicVI}, and Planetary Rover Path Planning tasks\cite{Pflueger2018RoverIRLIR}, VIN has been proven to be successful.

\begin{figure}[!htbp]
  \centering
  \includegraphics[width=0.45\textwidth]{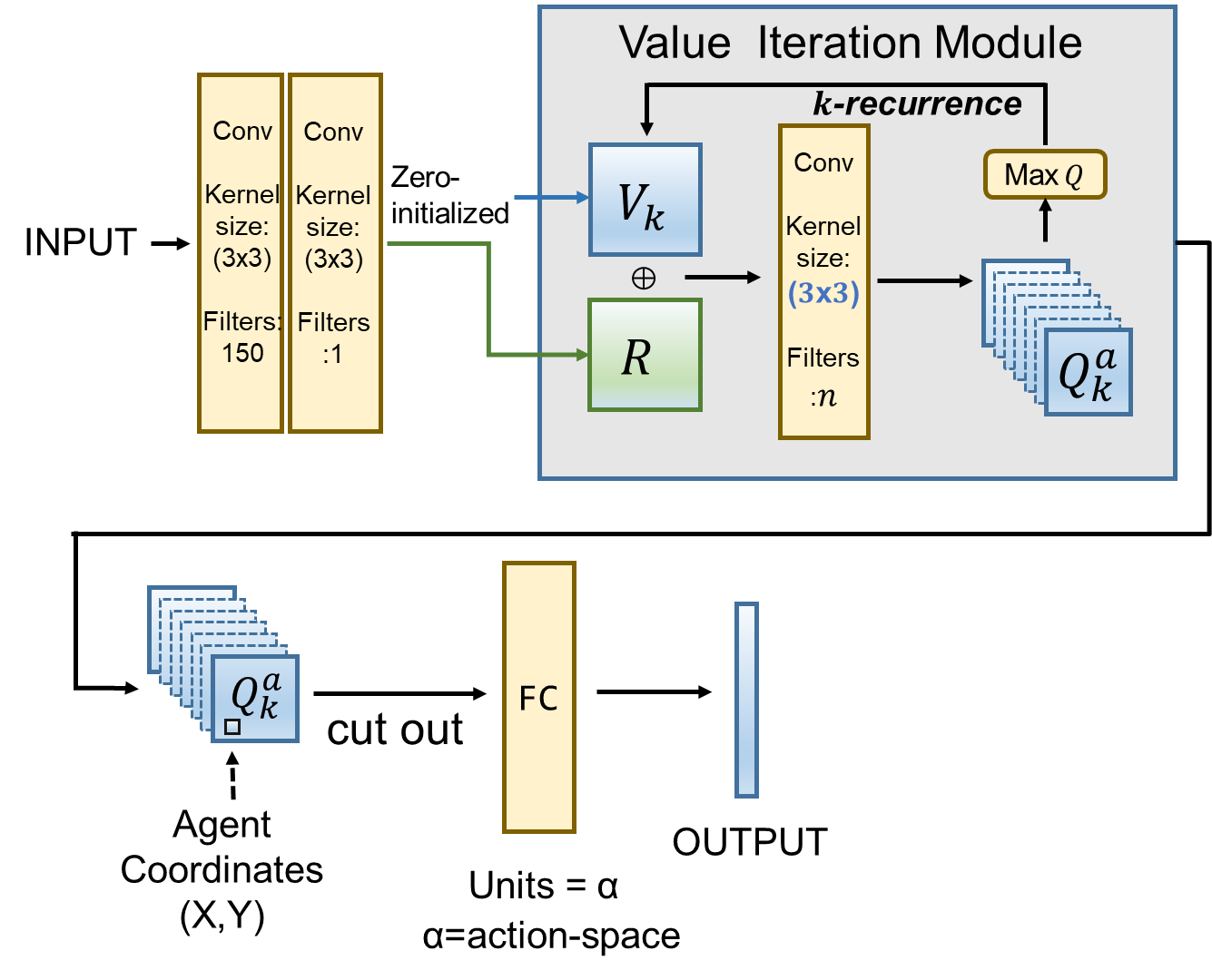}
  \caption{Overview of Value Iteration Network}
  \label{fig:VIN}
\end{figure}

The specific network structure of VIN is shown in Fig.~\ref{fig:VIN}. 
The input to the network is the observed map information $o$. 
VIN first goes through two CNN layers to generate a reward function $R$. 
Taking the path-finding task as an example, in this scenario, the reward function, value function, and action value function would all be instantiated as reward map $R$, value map $V$, and action-value map $Q^a$ with the same size as the input. 
Specifically, the position of the endpoint on the reward map tends to be a positive value, obstacles tend to be a negative value, and movable empty spaces tend to be 0.
The reward map and a value map initialized with 0 of the same size are input into the value iteration module. 
In the value iteration, the reward map and the value map are stacked and input into the convolutional layer, where the convolutional kernel size is $3 \times 3$ and the number of kernels is $n$, which is the encoded action-space. 
This calculates the impact of each action on the value of the map, resulting in the calculation of the action-value map, denoted as $Q^a$.
Next, the tensor $Q^a$ output by the convolutional layer is max-pooled along the action dimension. 

The max-pooled tensor is then used as the new value map, stacked with the reward map, and iterated again. This operation is repeated $k$ times until the values of each position on the value map converge. 
The tensor $Q^a$ after convergence is then extracted based on the input location information to obtain the value $Q(s,a)$ corresponding to the agent's position, and the fully connected (FC) layer decodes the extracted value to obtain the value of each action.

\subsection{Long Short-Term Memory}
Long Short-Term Memory (LSTM)\cite{LSTM} is a popular type of recurrent neural network (RNN)\cite{RNN} that is specifically designed to handle the vanishing gradient problem. The key idea behind LSTM is the use of gating mechanisms to selectively regulate the flow of information through the network. These gates are learned during training and allow the network to remember or forget information over time, depending on the relevance of that information for the current task.

The basic LSTM unit consists of a cell state $c_t$, an input gate $i_t$, a forget gate $f_t$, an output gate $o_t$, and a hidden state $h_t$. The cell state represents the internal memory of the unit and is updated through a combination of the input and forget gates. The input gate controls the flow of new information into the cell state, while the forget gate controls the flow of information out of the cell state. The output gate then determines how much of the cell state should be exposed to the rest of the network through the hidden state.

\begin{equation}\label{eq:lstm}
\begin{aligned}
f_t &= \sigma(W_f x_t + U_f h_{t-1} + b_f)\\
i_t &= \sigma(W_i x_t + U_i h_{t-1} + b_i)\\
\tilde{c}_t &= \tanh(W_c x_t + U_c h_{t-1} + b_c)\\
c_t &= f_t \odot c_{t-1} + i_t \odot \tilde{c}_t\\
o_t &= \sigma(W_o x_t + U_o h_{t-1} + b_o)\\
h_t &= o_t \odot \tanh(c_t)
\end{aligned}
\end{equation}

In the above equations, the variables are defined as follows:
\begin{itemize}
    \item $\sigma$: The sigmoid activation function, which is used to squash the input values between 0 and 1.
    \item$\odot$: The element-wise multiplication operator, which denotes the multiplication of corresponding elements in the involved matrices or vectors.
    \item $W_f$, $W_i$, $W_c$, $W_o$: The learned weight matrices associated with the input vector $x_t$ for the forget gate, input gate, cell update, and output gate, respectively.
    \item $U_f$, $U_i$, $U_c$, $U_o$: The learned weight matrices associated with the previous hidden state $h_{t-1}$ for the forget gate, input gate, cell update, and output gate, respectively.
    \item $b_f$, $b_i$, $b_c$, $b_o$: The learned bias parameters for the forget gate, input gate, cell update, and output gate, respectively.
    \item $x_t$: The input to the LSTM unit at time step $t$.
    \item $f_t$: The forget gate activation at time step $t$, determining the extent to which the cell state from the previous time step is retained.
    \item $i_t$: The input gate activation at time step $t$, controlling the extent to which new information is added to the cell state.
    \item $\tilde{c}_t$: The candidate cell state at time step $t$, representing the new information to be potentially added to the cell state.
    \item $c_t$: The updated cell state at time step $t$, taking into account both the retention of information from the previous cell state and the addition of new information.
    \item $o_t$: The output gate activation at time step $t$, controlling the extent to which the updated cell state influences the hidden state.
    \item $h_t$: The updated hidden state at time step $t$, which is used for subsequent time steps and serves as the output of the LSTM unit at the current time step.
\end{itemize}

LSTM has become a popular choice for tasks that involve sequential data, such as speech recognition\cite{graves2006connectionist,amodei2016deep}, language modeling\cite{sundermeyer2012lstm,radford2018improving}, and video analysis\cite{donahue2015long,srivastava2015unsupervised}, due to its ability to handle long-term dependencies and avoid the vanishing gradient problem.

\subsection{Convolutional LSTM}
Convolutional LSTM(ConvLSTM)\cite{ConvLSTM} is a variant of the LSTM recurrent neural network that is designed to handle sequences of multidimensional inputs, such as images or videos. 
It is a combination of Convolutional Neural Networks (CNNs) and LSTMs, where CNNs are used for feature extraction and LSTMs are used for handling sequential information.

The ConvLSTM cell has a convolutional structure inside the LSTM cell, which allows it to extract features from the input sequence and use them to update its hidden state. 
This structure also allows ConvLSTM to learn spatial hierarchies of features.

In ConvLSTM, the input gate, forget gate, and output gate are replaced by convolutional layers that perform the same functions as their traditional counterparts, but with the added ability to capture spatial correlations between the input data. 
Specifically, the computation in a ConvLSTM unit involves replacing the matrix multiplication in \eqref{eq:lstm} with convolutional operations.
This means that ConvLSTM can learn both temporal and spatial dependencies in the input sequences, and use them to make predictions.

ConvLSTM has demonstrated success in numerous computer vision applications, including action recognition, semantic segmentation, and video prediction.
They have also been used in other applications where multi-dimensional sequential data is involved, such as meteorology and oceanography.

\section{Related Work}\label{sec: Realted work}

The performance of Value Iteration Networks (VIN) has been the subject of several studies, aiming to improve their effectiveness in various aspects, such as reducing error accumulation, enhancing expressiveness, and handling larger input maps. In this section, we provide an overview of the main contributions in this field, organized by the nature of the improvements they propose.

The Value Iteration Residual Network (VIRN)\cite{VIRN} introduces two primary enhancements: Firstly, it employs larger convolutional kernels and shorter iterations within the VI module to reduce the module's depth. 
Secondly, it outputs the value functions from each iteration in the VI module to an attention-based summarization module, which then takes a weighted average based on learned weights to produce a summarized value function. This effectively summarizes the outcomes of each iteration in the VI module and reduces error accumulation. Experiments conducted in the Atari Pac-man game environment demonstrate that VIRN outperforms the VIN in terms of learning efficiency and solution quality.
Although the attention module in VIRN has been proven to possess some summarization capability, we believe that its capacity is limited since the structure relies on a single map unit and utilizes a simple linear combination. Furthermore, the specific influence of kernel size and iteration count in the VI module was not explored in \cite{VIRN}.
This study further investigates the relationship between the convolutional kernel size and the number of iterations within the VI module. In addition, we employ a more expressive ConvLSTM-based module as a new summarization module to further reduce error and enhance network accuracy.

Value Iteration Networks with Double Estimator (dVIN) \cite{DVIN} focuses on minimizing the error in each iteration. Xiang Jin et al. decouple action selection and value estimation in the VI module and employ a weighted double estimator method to approximate the maximum expected value, as opposed to maximizing the action value estimated by a single network. This approach effectively reduces the single-iteration error. Additionally, they design a two-stage training strategy to address the high computational cost and poor performance of the VI-based model in large-scale domains. Empirical studies on grid world domain and lunar terrain images indicate that dVIN outperforms baseline methods and is more suitable for large-scale environments.

Value Iteration Networks on Multiple Levels of Abstraction (AVIN) \cite{AVIN} also focuses on differentiating between local and global information. Unlike our work, AVIN divides the input map into several abstract levels for computation at the input stage. To compensate for the information loss caused by reduced resolution, AVIN increases the number of feature representations. The results calculated based on different abstraction levels are aggregated to obtain the action value for each action. AVIN is capable of solving larger 2D grid world planning tasks compared to the original VIN implementation and is successfully applied to omni-directional driving planning for search and rescue robots in cluttered terrain.


Jiang Zhang et al. \cite{DB-CNN} also focused on differentiating between local and global information. They proposed a novel deep convolutional neural network with double branches (DB-CNN) to address the problem of training difficulty caused by a large number of iterations in VIN. DB-CNN performs planning through both local and global feature extraction. The global feature extraction process is similar to the computation in the VI module of VIN. However, unlike the VI module in VIN, which actively performs global planning, DB-CNN uses a designed fixed structure for planning. Overall, the structure of DB-CNN is more inclined toward traditional reflexive networks.

\section{Method}\label{sec: Method}
In the VI module, the model needs to solve the optimal value function $V(s)$ by continuously iterating calculations. 
Each iteration step can be viewed as expanding local information and ultimately outputting a value function predicted based on global information. 
In fact, such an iterative mechanism may cause single calculation errors to accumulate continuously, and because of the existence of the maximum operation in the VI module, this error eventually tends to overestimate. 

To address this issue, GS-VIN applies two improvements: 1. using an adaptive iteration strategy in the VI module to reduce the number of iterations; 2. providing a gated summarization module to enable the network to have strong summarization capability to reduce the impact of accumulated errors during the iteration process.

\subsection{Adaptive iteration strategy}\label{par:ais}
Excessive iterations not only lead to error accumulation, but also increase the depth of the network, which can cause instability in training. 
The VI module cyclically convolves the value function and reward function through a convolutional layer and a special max-pooling layer, which is similar to the concept of RNN\cite{RNN}. 
Previous experience\cite{bengio1994learning} has shown that as the number of iterations in RNN increases, the network depth also increases, which can cause the network to become unstable during backpropagation due to gradient vanishing or exploding.
In VIN, the number of iterations in the VI module depends on the size of the network input, where a larger input requires more iterations so that the value of any point can propagate throughout the entire map for planning. 
This characteristic makes VIN difficult to handle larger inputs because it is more prone to gradient explosion and vanishing. 

In \cite{Ding2022ScalingUY}, the authors proposed a new convolutional neural network structure that uses large-sized convolutional kernels to effectively improve the accuracy of image classification tasks and has fewer parameters and higher computational efficiency than traditional small convolutional kernels. 
This study inspires us about the potential of using large convolutional kernels.
In the VI module, the convolutional layer is not used to extract features from the input, but to learn the state transition probabilities $P$. 
We speculate that the difference in the nature of this task may also lead to different effects of the convolutional kernel size compared to traditional tasks. 

However, for deep neural networks, blindly increasing the number of trainable parameters can make the network more difficult to train. 
Therefore, the relationship between the iteration times $k$ and the convolutional kernel size $f$, which directly affect the network depth, is particularly important in the VI module.

The essence of the VI module is to perform correct global planning on the input information. 
From this perspective, using larger convolution kernels in the VI module can increase the network's expressive power and improve planning speed.
However, using larger convolution kernels in the VI module is not harmless, as it can lead to training instability due to the increase in parameter size and network depth. 
In this case, we can reduce the number of iteration times $k$, to reduce network depth. 
In VIN, the number of iteration times $k$, is a manually selected hyperparameter passed into the network. 
We believe that to achieve a certain level of accuracy, the value of $k$ should be at least sufficient to enable the VI module to perform correct global planning on the input information. 
Specifically, in the VI module, the value at any point on the map should propagate to all points on the map at least once. 
Based on this conjecture, we have derived a relationship between the number of iterations $k$, the size of the input map $m \times n$, and the size of the convolution kernel $f$ in the VI module. 
The relationship is given by:

\begin{equation}\label{eq:mfk}
    k = \frac{\sqrt{m^2+n^2}}{\frac{f-1}{2}}
\end{equation}

where $\sqrt{m^2+n^2}$ is the maximum distance between any two points in the input map of size $m \times n$, and $(f-1)/2$ represents the radius of value propagation for a single iteration with a convolution kernel of size $f$. For $k$ we ceiling to the nearest integer.

Additionally, we will delve into a more detailed analysis in Section \ref{par:mkf}, where a multitude of experimental results will be presented to further validate our claims.
\subsection{Gated Summarization Module}
The iteration process in the VI module is accompanied by the accumulation of errors. 
This is because in each iteration step, the network model needs to use the learned transition probability matrix $P$ to estimate the value function of all possible action successor states. 
However, in neural networks, the learned transition probability matrix $P$ inevitably has errors compared to the true transition probability matrix $\overline{P}$, and the iterative calculation makes the error accumulate exponentially. 
Specifically, when the CNN in the VI module calculates the action-value function for all actions, the calculated result  has a certain error. 
The subsequent maximum operation uses the higher part of the error as the estimate for the value function, and the network model continues to expand exploration based on this overestimated value in the later iteration steps. 
This may cause us to ignore better estimates and ultimately lead to overestimation problems. 
This self-bootstrapping overestimation problem is quite common in value iteration algorithms\cite{thrun1993issues,DDQN}, and similarly, using neural networks to approximate the value iteration algorithm in the VI module also has this issue.


The iterative process in the VI module is a planning process that expands from local to global. 
As a result, global planning is necessary, but we believe that the process of global planning is equally important. 
For example, in reality, humans do not make decisions based solely on long-term predicted outcomes, but rather take into account both short-term and long-term predictions when taking action.
This is due to humans' limited prediction capability, and thus the predicted future may not match the actual future. 
As the time step increases, the error also increases, and there is a greater possibility of misjudging the current state. For example, in a game scenario where a player is faced with a high-risk trap, the best course of action is to retreat and look for other ways out, regardless of whether there is a tempting treasure after the trap. 
In this example, too much consideration of the action to obtain the treasure after the trap would only disrupt the assessment of the current situation.


The example above illustrates the importance of considering the process of future projection when making decisions, rather than solely focusing on the ultimate future outcome. 
Therefore, we output the results of each iteration of the VI module separately and use a new module to summarize them.

\begin{figure}[htbp]
	\centering
	\includegraphics[width=1\linewidth]{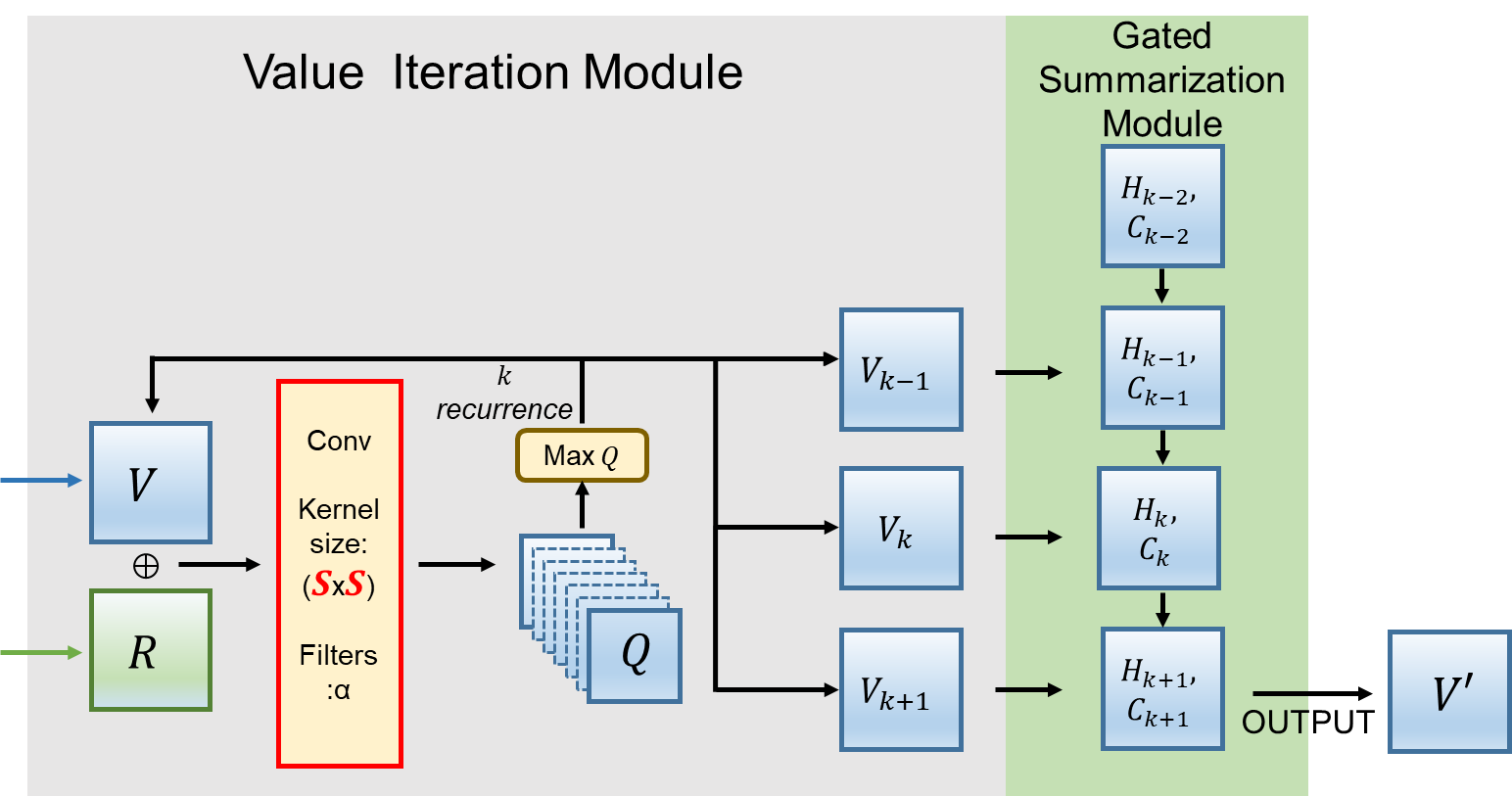}
	\caption{The structure between the GS module and the VI module.}
        \label{fig:lstmvin}
\end{figure}

In this study, we propose a novel gated summarization module that performs a temporal and spatial resampling of the entire planning process in the VI module. 
The new module employs a gating mechanism to achieve an effect similar to that of the attention mechanism. 
The fundamental computational unit is the value at each coordinate, and a LeakyReLU activation function is incorporated into the computation.
Ultimately, our module can be approximated as a special convolutional LSTM layer defined by \eqref{eq:ConVLSTM-VIN}:

\begin{equation}\label{eq:ConVLSTM-VIN} 
\begin{aligned}
&V_{k}=\max_{\alpha} (W_{\alpha}^{R} *R+ W_{\alpha}^{V} *V_{k-1})\\
&f_k = \sigma(W_f * V_{k} + U_f * h_{k-1}) \\
&i_k = \sigma(W_i * V_{k} + U_i * h_{k-1}) \\
&\tilde{c_k} = \text{LeakyReLU}(W_c * V_{k} + U_c * h_{k-1}) \\
&c_k = f_k \odot c_{k-1} + i_k \odot \tilde{c_k} \\
&o_k = \sigma(W_o * V_{k} + U_o * h_{k-1})\\
&h_k = o_k \odot \text{LeakyReLU}(c_k) \\
\end{aligned}
\end{equation}
As before, here ‘$*$’ denotes the convolution operator.
The reward function, denoted by $R$, is derived from the sampled observed information. 
The value function at the k-th iteration is represented by $V_k$. 
In the VI module's convolutional layer, the weights for the value function and reward function are given by $W^{V}$ and $W^{R}$, respectively.
$i_k$, $f_k$, and $o_k$ represent the input gate, forget gate, and output gate, respectively. These gating units compute the gate values based on the input $V_k$ at the current time step and the hidden state $h_{k-1}$ from the previous time step, thereby regulating the flow of information between cells.
Additionally, the action index is denoted by $\alpha$.

LSTM can be viewed as an attention mechanism with gating mechanisms. 
It adaptively selects and stores important information based on the input sequence and outputs it when needed. 
The use of convolutional operations enables the network to maintain spatial information. 
In GS-VIN, bias information is unnecessary, and linear activation functions should be used for the convolutional operations of the value iteration operation to ensure the effective preservation of the value function information. However, due to the use of GS module, the network complexity is almost doubled. 
In order to maximally retain the original information of the value function while minimizing the impact of vanishing and exploding gradients, we choose to use LeakyReLU as the activation function instead of linear or Tanh activation functions.



Fig.~\ref{fig:lstmvin} shows the architecture between the VI module and the GS module.
In the GS-VIN model, the value function at each iteration step within the VI module is stacked along a new temporal axis. 
After completing all iterations, the VI module output a new tensor containing the results of the value function for each iteration step. 
This tensor can be approximately interpreted as representing the entire planning process from local to global in the VI module. 
The GS module follows the VI module.
Within the GS module, the network continuously performs convolutional operations on the value function along the temporal axis of the input tensor to sample spatial information. 
Based on this spatial information and its gating structure, the network determines the current internal state $h_k$ as well as whether to reinforce or weaken the memory cell state $c_k$. 
Finally, the GS module outputs the computed internal state $h_k$ as the summarized value function $V'$ for subsequent operations.

\section{Evaluation and Discusstion}\label{sec: Evaluation}
\begin{figure}[!htbp]
	\centering
	\begin{minipage}{0.49\linewidth}
		\centering
		\includegraphics[width=0.9\linewidth]{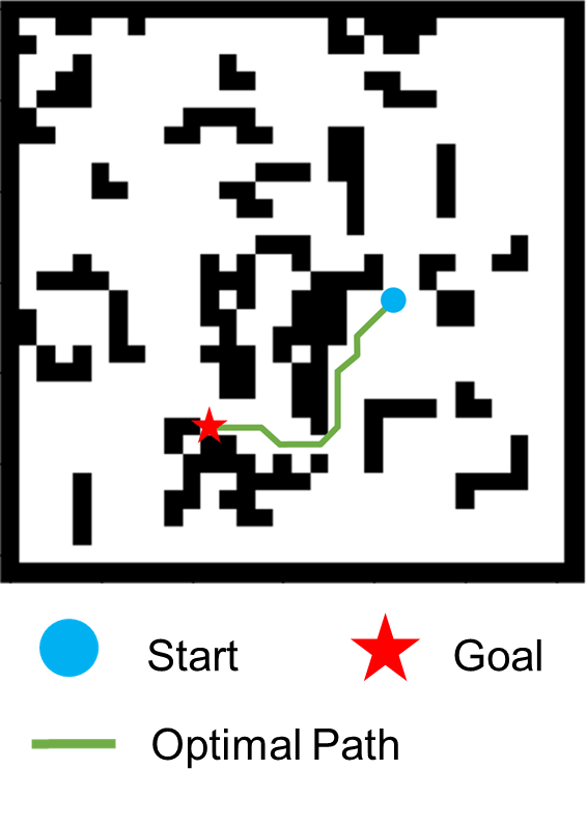}
	\end{minipage}
	\begin{minipage}{0.49\linewidth}
		\centering
		\includegraphics[width=0.9\linewidth]{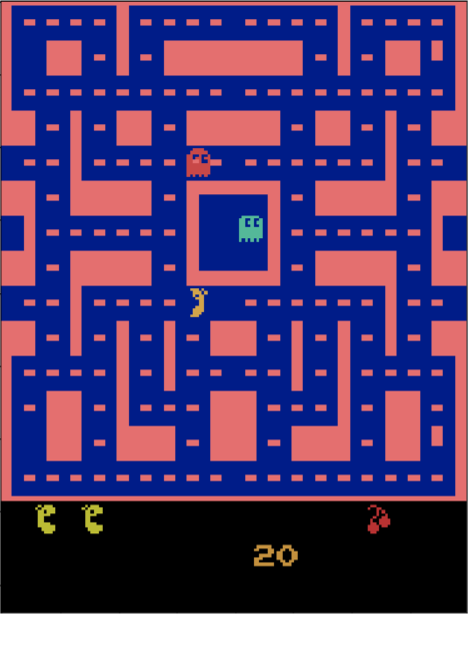}
	\end{minipage}
    \caption{Random instances of 2D Grid-World Domain and Mr. Pac-Man Domain}
    \label{fig:example}
\end{figure}

In this section, we conduct experiments and evaluations of GS-VIN in two different domains: the 2D Grid-World Domain and the Mr. Pac-Man Domain. An illustrative example of these environments can be seen in Fig.~\ref{fig:example}. The primary goal of these experiments is to verify whether GS-VIN can effectively perform global planning across diverse domains, and further assess the impact of the two proposed improvements: the adaptive iteration strategy and the GS module. Finally, we introduce a heuristic function for the adaptive iteration strategy, validate its effectiveness through experiments, and discuss the implications of this heuristic function for VI-based models.

\subsection{2D Grid-World Domain}\label{par:2dmain}
Our first experimental domain is a classic 2D grid-world path-finding problem. 
The 2D Grid-World path-finding problem is a task that involves identifying the shortest route between a start point and an endpoint(goal) within a discrete two-dimensional grid space. 
In this setting, the grid typically consists of a start point, an endpoint, and several obstacles, such as walls. The objective of an agent navigating this environment is to determine the most efficient path from the start to the target while circumventing any obstacles present.
In 2D Grid-World domain, we can effectively test the fundamental path-planning ability of different models, which helps us better understand and study the differences between algorithms.

\subsubsection{Experimental Setting}
In this experimental domain, obstacles occupy a proportion of grid cells that is proportional to the size of the map within a certain range. 
The agent's goal is to find the shortest path from its current location to the goal, and it can only take action to move to one of the eight neighboring cells. 

The network takes as input two map information and agent coordinates, where the first map is a binary obstacle map with 0 for empty space and 1 for obstacles, and the second map is a goal map with 10 at the goal position and 0 elsewhere. 
The final map information has a size of $m \times n \times 2$, where $m$ and $n$ are the dimensions of the map, and the agent coordinates are added as additional input. 
The model outputs a probability distribution over actions for the agent. 
In the 2D grid-world path-finding domain, we use imitation learning to train the network model. 
We use the directions of the shortest path obtained by the A* algorithm as the training labels, convert them into one-hot vectors, and compare them with the network's output to compute the loss.

The training settings include 30 epochs, a batch size of 256, a learning rate of 0.002, and cross-entropy as the loss function. 
The dataset consists of 1,024,800 different maps, and each map randomly generates 6 sets of agent and goal coordinates. 
The dataset is split into a training set and a test set in an 8:2 ratio for evaluation. The main problem of VI-based network training is its instability.
Therefore, in the experiment, the weight matrices are initialized with a standard normal distribution with a mean of 0 and a standard deviation of 0.01 to increase the stability of training.



In this section, we first evaluate the performance of VIN, VIRN, GS-VIN, and DB-CNN on classic path-finding problems. 
VIN, VIRN, and the proposed GS-VIN in this study are VI-based models, with their most prominent feature being the retention of the active planning module (i.e., the VI module). 
In contrast, DB-CNN\cite{DB-CNN} is a traditional reflexive neural network inspired by VIN, featuring a dual-branch structure but lacking an iterative module.

In Section~\ref{par:classic}, the evaluation metrics include Accuracy, Success Rate, and Trajectory Difference. 
Accuracy refers to the percentage of cases in which the agent selects the same action as the ground truth label out of the 8 possible actions at a given state, i.e., single-step accuracy. 
Success Rate indicates the proportion of times the agent successfully reaches the goal from the initial position in a given map. 
Trajectory Difference denotes the difference in the length of the actual path and the true shortest path in instances where successful planning to the endpoint occurs.
It is important to note that in Section~\ref{par:classic}, each model was set to use the recommended hyperparameters specified by the respective authors.
Specifically, VIRN and GS-VIN used a convolutional kernel size $f$ of 11, with 5 iterations for $16 \times 16$ maps, 10 iterations for $32 \times 32$ maps, and 19 iterations for $64 \times 64$ maps. 
For VIN, the paper did not provide recommended iteration numbers for maps of size 32 and 64, so we inferred them from the reference values given in \cite{VIN}. 
Finally, VIN used a convolutional kernel size $f$ of 3, with 20 iterations for $16 \times 16$ maps, 43 iterations for $32 \times 32$ maps, and 87 iterations for $64 \times 64$ maps.
We omit the explanation of these hyperparameters for DB-CNN as it does not have a VI module.

In Section~\ref{par:mkf}, we conduct experiments and analyze the specific application of the adaptive iteration strategy proposed in this study. 
By carrying out extensive experiments on the iteration count $k$ and convolution kernel size f of the VI module in VI-based models, we aim to provide a reference heuristic function. 
In this section, the non-VI-based model DB-CNN is not experimented with, and the comparison object in the experiments is the single-step accuracy, which has a more direct relationship with network performance.

\subsubsection{Performance on 2D Grid-World Domain}\label{par:classic}

\begin{table}[ht]
\centering
\caption{Performance on different domain sizes.}
\begin{tabular}{cllll}
\hline
\multicolumn{1}{l}{Domain Size} & Methods & Accuracy & Succ.rate & Traj.Diff \\ \hline
\multirow{3}{*}{16x16}          & VIN     & 0.8914   & 0.9144    & \textbf{0.00191}   \\
                                & VIRN    & 0.9202   & 0.9414    & 0.00486   \\
                                & DB-CNN    & \textbf{0.9746}   & \textbf{0.9919}    & 0.00290   \\
                                & GS-VIN & 0.9605   & 0.9769    & 0.00531   \\ \hline
\multirow{3}{*}{32x32}          & VIN     & 0.8797   & 0.8749    & \textbf{0.00181}   \\
                                & VIRN    & 0.8986   & 0.8829    & 0.00858   \\
                                & DB-CNN    & 0.8987   & 0.9559    & 0.00466   \\
                                & GS-VIN & \textbf{0.9417}   & \textbf{0.9563}    & 0.00254   \\ \hline
\multirow{3}{*}{64x64}          & VIN     & 0.5365   & 0.2186    & 0.04476   \\
                                & VIRN    & 0.8232   & 0.6743    & 0.00811   \\
                                & DB-CNN    & 0.6464   & 0.4247    & 0.24713   \\
                                & GS-VIN & \textbf{0.8946}   & \textbf{0.7463}    & \textbf{0.00542}   \\ \hline
\end{tabular}
\label{tab:TP1}
\end{table}


\begin{figure*}[!htbp]
	\centering
	\begin{minipage}{0.51\linewidth}
		\centering
	    \includegraphics[width=1\linewidth]{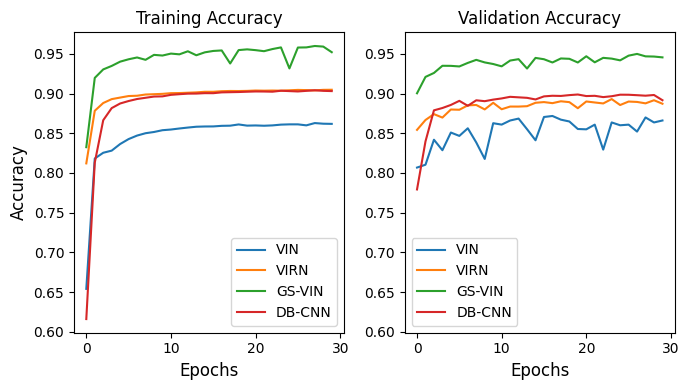}
	    \caption{Comparison of different models in 32$\times$32 mapsize}
            \label{fig:TP1}
	\end{minipage}
	\begin{minipage}{0.48\linewidth}
		\centering
		\includegraphics[width=1\linewidth]{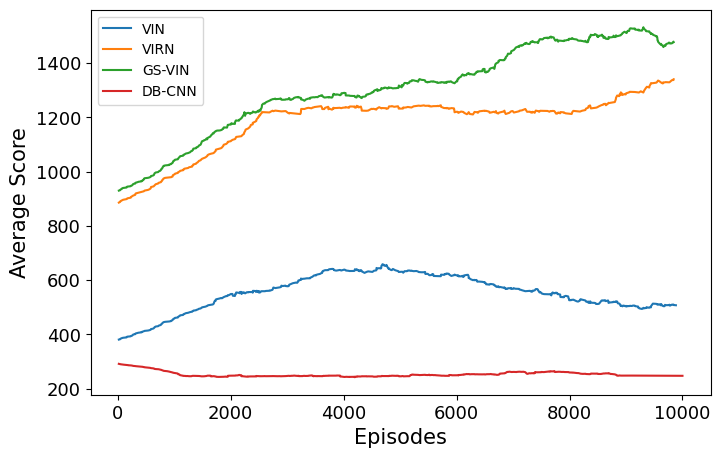}	
		\caption{Performance comparison of VIN, VIRN, GS-VIN, and DB-CNN in Mr. Pac-Man}
            \label{fig:pacman}
	\end{minipage}
\end{figure*}

Table~\ref{tab:TP1} shows the experimental results of VIN, VIRN,DB-CNN, and GS-VIN based on different map sizes. 

For the $16\times16$ grid world, the GS-VIN method achieves an accuracy of 0.9605 and a success rate of 0.9769, significantly higher than those of VIN and VIRN. 
However, DB-CNN boasts the best accuracy and success rate of 0.9746 and 0.9919, respectively. 
In terms of trajectory difference, VIN performs the best, while GS-VIN obtains the worst results. 
For the $32\times32$ grid world, the results are slightly different; GS-VIN has an absolute leading accuracy of 0.9417 and a slightly higher success rate of 0.9563 compared to DB-CNN. 
In trajectory difference, VIN still performs exceptionally well, followed by GS-VIN, while DB-CNN and VIRN fare poorly. 
For the $64\times64$ grid world, VIN and DB-CNN lag in accuracy; regarding success rate, it is worth noting that the average path length in $64\times64$ maps is longer, requiring more planning steps in a single path-finding task, hence the accumulative effect of a single-step accuracy increases. 
Ultimately, VIN and DB-CNN achieve success rates of 0.2186 and 0.4247, respectively, while GS-VIN maintains relatively excellent results with an accuracy of 0.8946 and a success rate of 0.7463. 
In trajectory difference, GS-VIN performs best, but we notice a more significant contrast in the results for VIN, especially when compared to smaller domain sizes.

The trajectory difference is defined as the difference between the actual path length and the true shortest path length in the trajectories that successfully reach the endpoint.
In theory, higher single-step accuracy in successful planning paths should result in a smaller trajectory difference. 
However, in the comparison between GS-VIN and VIN in Table~\ref{tab:TP1}, especially for the $16 \times 16$ and $32 \times 32$ map sizes, GS-VIN exhibits higher single-step accuracy but significantly worse trajectory difference than VIN. 
We analyze that failed single-step predictions in VIN lead to more severe consequences, such as moving toward obstacles. 
In other words, if VIN makes an erroneous single-step planning, there is a higher probability that the entire path-finding task fail, ultimately resulting in fewer erroneous single-step predictions in successful path-finding examples. 
In contrast, GS-VIN is less likely to cause severe consequences in erroneous single-step predictions, resulting in more incorrect single-step predictions in successful path-finding examples and a larger trajectory difference.
It should be noted that although GS-VIN's performance in terms of path differences is not as good as that of other comparison subjects, its worst score still only exhibits a path difference of 0.00542. In absolute terms, this remains a very small value.

Overall, GS-VIN demonstrates the best comprehensive performance. 
For DB-CNN, in domain sizes of $32\times32$ and below, it exhibits performance similar to or even better than GS-VIN. 
However, in the larger $64\times64$ domain size, there is a substantial contrast, with its performance only better than VIN. 
We believe this is mainly because DB-CNN is a non-iterative fixed-depth neural network. 
When the input size increases, DB-CNN cannot increase the iteration count like VI-based models to accomplish planning for larger sizes. 
Insufficient processing of input information causes DB-CNN's performance to decline rapidly. VIN's performance also drops sharply in larger domain sizes, which we attribute to its excessive iteration count, leading to the inevitable accumulation of single-step iteration errors and poor performance. 
GS-VIN and VIRN share the same convolution kernel size and iteration count, with the performance difference between the two resulting from the impact of different summarization module.

Fig.~\ref{fig:TP1} illustrates the learning curve when the map size is $32 \times 32$.
From the figure, we can infer that GS-VIN has the highest learning efficiency, achieving the highest accuracy under the same number of epochs. 
However, compared to other models, GS-VIN's training is relatively more unstable, which we believe is mainly due to the increased network depth caused by the GS module. 
In conclusion, through the above analysis, we can deduce that, owing to the stronger summarization capability of the GS module, GS-VIN exhibits the highest single-step accuracy and path-finding success rate, significantly outperforming VIN. 
It is worth noting that the gap between the data increases with the map size.

\subsection{Mr.Pac-Man Domain}

In this section, we aim to evaluate the performance of our improved value iteration network model in more complex environments beyond the traditional 2D grid-world path-finding problems. 
We employ the classic game Mr.Pac-Man as a test domain, as it shares the fundamental nature of path-finding tasks but exhibits several additional complexities:
\begin{itemize}
\item Dynamic environment: In contrast to static path-finding tasks, the Mr.Pac-Man game environment is more dynamic. Ghosts move around the map, making the path-finding challenge even more demanding.

\item Real-time decision-making: Agents in the Pac-Man game must make decisions in real time. Certain in-game events may fundamentally alter the strategy an agent should adopt. 
For example, after obtaining special pellets from the map's corners, invulnerable enemy ghosts become score-earning units for a limited time. 
In this bonus time, the agent's strategy should shift from avoiding ghosts to moving toward them and consuming them to maximize the score. 
This requires the network model to adjust in real time according to environmental changes. 
Under these circumstances, the model's reinforcement learning capabilities may prove to be even more prominent.

\item Multi-objective optimization: Unlike simple path-finding tasks, Pac-Man gameplay requires balancing multiple objectives, such as collecting pellets while avoiding ghosts. This adds further complexity to the problem.

\item State-space complexity: The state space of the Pac-Man game is more intricate compared to traditional path-finding tasks, as it must account for not only the map information but also the status of ghosts and pellets. This necessitates the model to learn and make decisions within a larger state space. Although this increases the computational burden, it also helps validate the model's performance in complex environments.
\end{itemize}

\subsubsection{Experimental Setting}
These environmental characteristics can help us better evaluate the capability of different network models in handling complex tasks. 
For the specific training approach, we reference the method outlined in \cite{DQN}. 
The map is compressed into an $84\times84\times1$ grayscale image, and in order to capture the velocity and directional information within the game screen, we use groups of 4 frames, resulting in a final network input size of 84x84x4. 
The reward allocation in the environment is based on the game score. 
Deep Q-learning was employed as the baseline training method, with a learning rate of 0.0001, reward decay of 0.99, the epsilon-greedy value of 0.05, and the target network updated every episode. 
Mini-batch optimization was also utilized, with a batch size of 32. The loss function was defined as per \eqref{eq:loss}.


\begin{equation}\label{eq:loss}
\begin{aligned}
    &targetQ=r+\gamma*Qmax_{a'}(s',a')\\
    &loss=E[(targetQ-Q(s_t,a_t))]^2
\end{aligned}
\end{equation}

We compare the performance of VIN, VIRN, DB-CNN, and GS-VIN in terms of their game scores. 
In the VIN model, the VI module utilizes a convolutional kernel with $f=3$, while in the VIRN and GS-VIN models, a convolutional kernel with $f=11$ is used. 
For the number of iterations k, we employ the calculation given in \eqref{eq:mfk} with 119, 24, and 18.

\subsubsection{Performance on Mr.Pac-Man Domain}

As can be seen from Fig.~\ref{fig:pacman}, GS-VIN not only achieves the highest score performance, but also consistently obtains the highest scores during the same episode, indicating that GS-VIN demonstrates excellent learning efficiency in complex reinforcement learning tasks. 
Overall, VIRN ranks second, while VIN and DB-CNN exhibit noticeably inferior performance. We believe that the main reason for these results is the same as the cause of the outcomes in the $64 \times 64$ scenario described in Section~\ref{par:classic}: for VIN, when the input size becomes too large, an excessive number of iterations leads to the accumulation of single-iteration errors, ultimately resulting in poor performance. 
In the case of DB-CNN, its fixed reflective network architecture prevents it from effectively handling larger input information.


\subsection{Relationship between Convolutional Kernel Size and Iteration Times in VI Module}\label{par:mkf}
In this section, we begin by discussing the necessity of utilizing the adaptive iteration strategy. We present a heuristic function \eqref{eq:mfk} in section \ref{par:ais} that illustrates the relationship between map size, VI module kernel size, and VI module iteration count. Through extensive experimentation, we evaluate the performance of different VI-based models and provide recommendations for utilizing the adaptive iteration strategy in VI-based models.
\subsubsection{Necessity of Adaptive Iteration Strategy}

\begin{table*}[!htbp]
\centering
\caption{Accuracy performance of VIN with different kernel sizes was evaluated under a $28\times28$ 2D grid-map when the iteration times $k$ was fixed at 36.}
\begin{tabular}{l|lllll}
\hline
         & kernel size=3 & kernel size=5 & kernel size=7 & kernel size=9 & kernel size=11 \\\hline
Accuracy & 0.890697      & 0.89508       & 0.892717      & 0.907755      & 0.899984       \\ \hline     
\end{tabular}
\label{tab:VIN_28}
\end{table*}

\begin{figure*}[ht]
	\centering
	\includegraphics[width=0.95\linewidth]{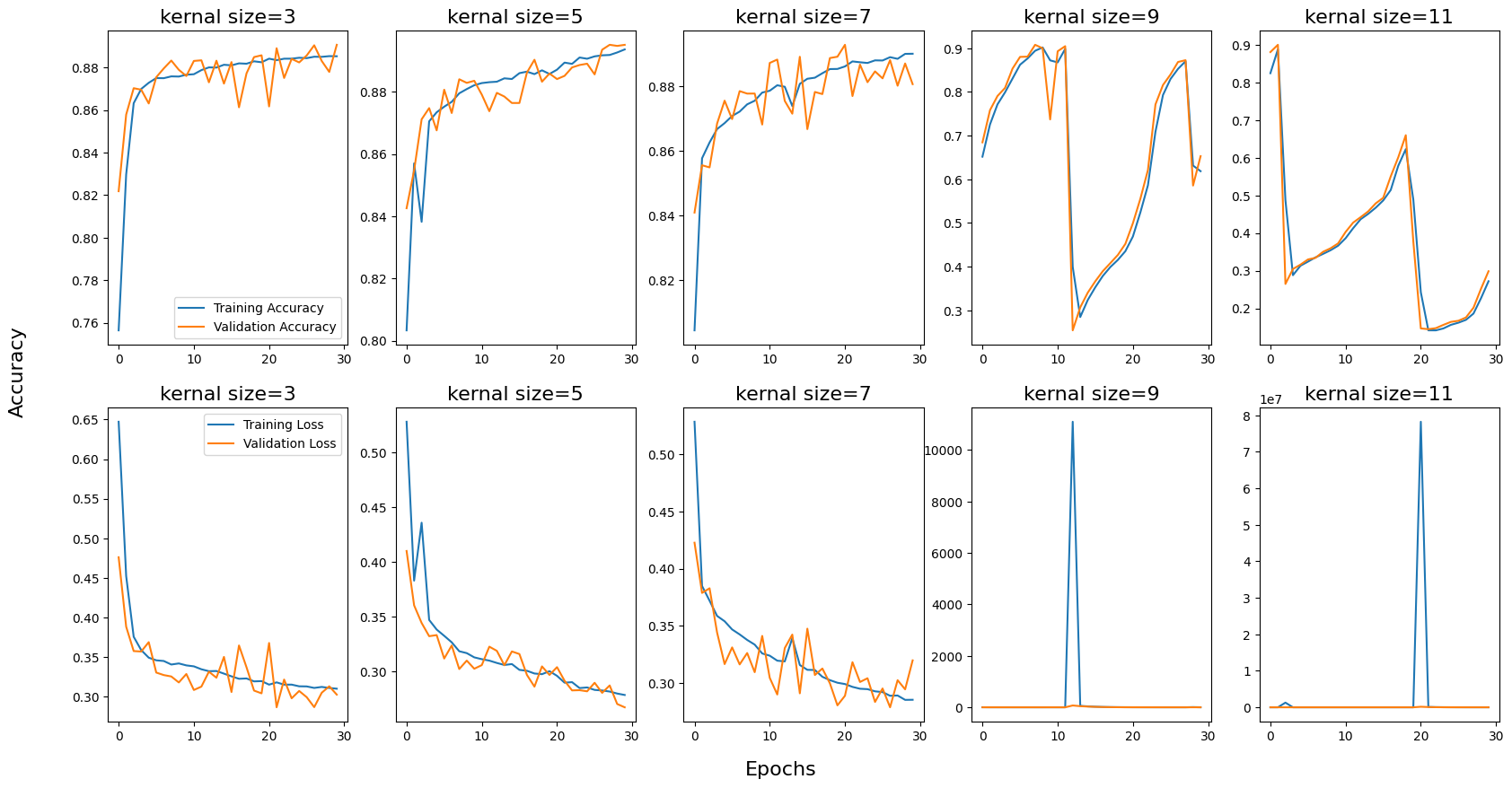}
	\caption{Training performance of VIN with different kernel sizes was evaluated under a $28\times28$  2D grid-map when the iteration times $k$ was fixed at 36. Top: training accuracy and validation accuracy. Bottom: training loss and validation loss.}
        \label{fig:VIN_28}
\end{figure*}
In the VI module of \cite{VIN}, the size of the convolutional kernel $f$ is fixed at 3, while the iteration time $k$ is set as hyperparameters, with recommended values of $k$ = 10 for 8$\times$8 domains, $k$ = 20 for 16 $\times$16 domains, and $k$ = 36 for 28 $\times$28 domains.

We conducted experiments on changing the convolutional kernel size $f$ in VIN, and Table~\ref{tab:VIN_28} shows the single-step accuracy performance of VIN under a fixed $k=36$ in a $28\times28$ map. 
We can see that the recommended $f=3$ in \cite{VIN} is not the optimal solution, and $f=9$ has higher single-step accuracy. 
However, it is worth noting that the training process begins to experience gradient explosion since $f=9$ as shown in Fig.~\ref{fig:VIN_28}. 
Overall, Table~\ref{tab:VIN_28} provides preliminary evidence that $f=3$ is not the optimal solution. From the perspective of enhancing network accuracy and reducing error accumulation, the adaptive iteration strategy is necessary. However, blindly increasing the size of the convolutional kernel may still lead to an exponential growth in the number of network parameters, resulting in training instability. In this context, the relationship between the size of the convolutional kernel and the number of iterations becomes of paramount importance.

In Section \ref{par:ais}, we propose a heuristic function \eqref{eq:mfk} to demonstrate a relationship between the number of iterations $k$, the size of the input $m \times n$, and the size of the convolution kernel $f$.
To validate the effectiveness of this relationship, we used it as a baseline and compared it to values $k'$ that were 0.5,0.75,1.25,1.5 and 2 times larger than the baseline. 
We set $f$ to 3, 5, 7, 9, 11, 13, and 15, and conducted experiments on VIN, VIRN, and GS-VIN using $32\times32$ maps. 
The results are shown in Table~\ref{tab:32_kf},\ref{tab:kfnumber},\ref{tab:kftime}.





\subsubsection{Performance on Different Convolutional Kernel Size and Iteration Times}

\begin{table*}[!htbp]
\centering
\caption{The impact of different convolution kernel sizes $f$ and different iteration number coefficient $k'$ on the accuracy in VIN,VIRN and GS-VIN. "*" represents that gradient explosion or gradient vanishing occurred during the training process, and the result is marked as invalid.}
\begin{tabular}{l|llllllll}
\hline
VIN     & $f$=3    & $f$=5    & $f$=7    & $f$=9    & $f$=11   & $f$=13   & $f$=15   \\ \hline
$k'$=0.5  & 0.8028 & 0.8675 & 0.8748 & 0.8815 & 0.8924 & 0.8847 & 0.9026 \\
$k'$=0.75 & 0.8183 & 0.8770 & 0.8914 & 0.9049 & 0.9084 & 0.9071 & 0.9086 \\
$k'$=1    & \textbf{0.8797}* &0.9313 & 0.9055 & 0.9194 & 0.9150 & 0.9098 & 0.9071 \\
$k'$=1.25 & 0.8665* & 0.8837 & \textbf{0.9262} & 0.9229 & 0.9121 & 0.9118 & 0.9061 \\
$k'$=1.5  & 0.8761* & \textbf{0.9355} & 0.9092 & \textbf{0.9253} & 0.9271 & \textbf{0.9136} & \textbf{0.9073} \\
$k'$=2.0  & 0.8514* & 0.9260 & 0.9063 & 0.9219 & \textbf{0.9308} & 0.8995 & 0.9060 \\ \hline

\hline
VIRN     & $f$=3    & $f$=5    & $f$=7    & $f$=9    & $f$=11   & $f$=13   & $f$=15   \\ \hline
$k'$=0.5  & 0.8570 & 0.8781 & 0.8898 & 0.8964 & 0.8976 & 0.8940 & 0.9069 \\
$k'$=0.75 & 0.8804 & 0.9024 & 0.9141 & 0.9143 & 0.9096 & 0.9114 & \textbf{0.9131} \\
$k'$=1    & \textbf{0.8977} & 0.9299 & \textbf{0.9324} & 0.9137 & 0.9015 & \textbf{0.9130} & 0.9105 \\
$k'$=1.25 & 0.8650* & 0.9333 & 0.9135 & 0.9097 & \textbf{0.9137} & 0.9062 & 0.9129 \\
$k'$=1.5  & 0.8634* & \textbf{0.9389} & 0.9230 & 0.9115 & 0.9081 & 0.9072 & 0.9039 \\
$k'$=2.0  & 0.8778* & 0.9285 & 0.9121 & \textbf{0.9148} & 0.9042 & 0.9064 & 0.9003 \\ \hline

\hline
GS-VIN  & $f$=3    & $f$=5    & $f$=7    & $f$=9    & $f$=11   & $f$=13   & $f$=15   \\ \hline
$k'$=0.5  & \textbf{0.8354}* & \textbf{0.9099}* & 0.9474 & 0.9566 & 0.9462 & 0.9403 & \textbf{0.9516} \\
$k'$=0.75 & 0.2427* & 0.9096* & \textbf{0.9673} & \textbf{0.9617} & 0.9351 & \textbf{0.9529} & 0.9210 \\
$k'$=1    & 0.2916* & 0.2265* & 0.9603 & 0.9148 & 0.9417 & 0.9220 & 0.0768*\\
$k'$=1.25 & 0.3514* & 0.2893* & 0.9568 & 0.9493 & \textbf{0.9604} & 0.0768* & 0.8411* \\
$k'$=1.5  & 0.2558* & 0.1129* & 0.8947* & 0.9427 & 0.9197 & 0.1697* & 0.9386 \\
$k'$=2.0  & 0.1358* & 0.2655* & 0.5756* & 0.6169* & 0.8136* & 0.6185* & 0.2325* \\ \hline
\end{tabular}
\label{tab:32_kf}
\end{table*}

\begin{table}[ht]
\caption{The actual number of iterations $k$ under a $32\times32$ map size}
\centering
\begin{tabular}{l|llllllll}
\hline
       & $f$=3    & $f$=5    & $f$=7    & $f$=9    & $f$=11   & $f$=13   & $f$=15   \\ \hline
$k'$=0.5  & 23 & 12 & 8  & 6  & 5  & 4  & 4  \\
$k'$=0.75 & 34 & 17 & 12 & 9  & 7  & 6  & 5  \\
$k'$=1    & 46 & 23 & 16 & 12 & 10 & 8  & 7  \\
$k'$=1.25 & 57 & 29 & 19 & 15 & 12 & 10 & 9  \\
$k'$=1.5  & 68 & 34 & 23 & 17 & 14 & 12 & 10 \\
$k'$=2.0  & 91 & 46 & 31 & 23 & 19 & 16 & 13 \\ \hline
\end{tabular}
\label{tab:kfnumber}
\end{table}

\begin{table}[ht]
\caption{The Percentage of Time Spent on Iterations $k'$ with Kernel Size $f$ Relative to $f=3, k'=1$ for a $32 \times 32$ Map in VIN. Using an Intel Xeon Processor W-2265, 256GB (64GB x 4) of Memory, 1TB SSD, and NVIDIA Geforce RTX 3090 24GB GPU}
\resizebox{\linewidth}{!}{
\centering
\begin{tabular}{l|llllllll}
\hline
VIN    & $f$=3    & $f$=5    & $f$=7    & $f$=9    & $f$=11   & $f$=13   & $f$=15   \\ \hline
$k'$=0.5  & 58\% & 49\% & 43\%  & 42\%  & 46\%  & 36\%  & 51\%  \\
$k'$=0.75 & 79\% & 62\% & 56\% & 55\%  & 58\%  & 47\%  & 48\%  \\
$k'$=1    & \textbf{100\%} & 81\% & 78\% & 76\% & 72\% & 57\%  & 73\%  \\
$k'$=1.25 & 122\% & 94\% & 80\% & 78\% & 87\% & 70\% & 78\%  \\
$k'$=1.5  & 139\% & 112\% & 107\% & 104\% & 94\% & 77\% & 98\% \\
$k'$=2.0  & 183\% & 120\% & 120\% & 115\% & 129\% & 101\% & 106\% \\ \hline
\end{tabular}
}
\label{tab:kftime}
\end{table}

In this section, we analyzed the performance of three VI-based models with different convolutional kernel size $f$ and iteration number $k$ to evaluate the effectiveness of the proposed improvements: the adaptive iteration strategy and the GS module. We compared the performance differences between the models, taking into account the impact of these factors.

Based on the data presented in Table~\ref{tab:32_kf}, it is evident that the adaptive iteration strategy showed a clear advantage in improving network accuracy and training stability compared to a fixed convolutional kernel size and iteration number. By varying f and k, the models achieved higher accuracy, and the risk of gradient vanishing or exploding was reduced. Furthermore, Table~\ref{tab:kftime} highlights the positive impact of the adaptive iteration strategy on computation time. In most cases, the strategy allowed the network to maintain equivalent or superior accuracy performance while reducing training time, effectively cutting down the time cost of training the model.

The GS module was another essential improvement in the GS-VIN model. When comparing the performance of the three models under the same convolutional kernel size and iteration number settings, GS-VIN consistently exhibited higher overall accuracy among valid results. This indicates that the GS module effectively enhances the network's planning accuracy.

In conclusion, the adaptive iteration strategy has been shown to improve both the accuracy and training stability of network models, while the GS module specifically enhances the accuracy of the models. The improvement in network accuracy indicates that there is a smaller error in the network's results. By employing these strategies, we can achieve better network performance in various settings. It is worth mentioning that, in this experiment, in this experiment, our primary goal was to compare the performance differences caused by the network architectures. Therefore, we did not address training stability issues in the training methods themselves.  Further research can be done to optimize training methods and techniques, such as gradient clipping, to further improve the stability and performance of these models.

\subsubsection{Discussion on the Application of Adaptive Iteration Strategy to VI-based Models}
In this section, we discuss the impact of different values of $k'$ and $f$ on the performance of VIN, VIRN, and GS-VIN models, as well as the effectiveness of the proposed heuristic function \eqref{eq:mfk} for adaptive iteration strategy. We explore the relationship between these parameters and network performance, training stability, and model complexity, aiming to provide valuable insights for researchers working with VI-based models.

Regarding the relationship between $k'$ and $f$, our experiments revealed that for VIN, the performance is significantly better when $k'=1$ or above compared to when it is below $k'=1$. 
This is consistent with our initial prediction: \emph{The value of $k$ should be at least sufficient to enable the VI module to perform correct global planning on the input information.} 
Specifically, in our experiments with VIN, the network exhibits optimal overall performance when $k'=1.5$. 
For VIRN, the required $k'$ value for achieving the highest accuracy tends to decrease overall. 
We believe this is because the VIRN structure is more complex than VIN, and the network output does not solely rely on the final iteration result of the VI module, thus reducing the dependency on the number of iterations $k$ to some extent. 
GS-VIN takes this a step further, as lower $k'$ values generally perform better than higher ones. 
We attribute this to the addition of the GS module, which doubles the actual network depth and also involves convolutional computations. 
These computations summarize the calculation process in the VI module while also taking part in global planning inference, ultimately reducing the dependency on the number of iterations in the VI module. 
Although there might be some discrepancies for different network models, we generally find that $k'=1$ often yields the best overall performance. 
A higher number of iterations may sometimes result in an accuracy improvement, but the increased network depth can also make the training process more unstable.

Regarding the choice of $f$, our experiments show that on a $32\times32$ map size, the model generally performs better when $f=5$ or $f=7$, although there are exceptions. 
For GS-VIN, since $f=5$ still requires a higher number of iterative calculations, the network training becomes highly unstable. 
Ultimately, $f=7$ yields the best accuracy performance for GS-VIN. 
As shown in Table~\ref{tab:kfnumber}, although an increase in $f$ can reduce the required number of iterations and subsequently the network depth, the increase in $f$ also causes the parameters in the convolutional layer to grow exponentially. 
Although this has little impact on VIN and VIRN, in complex network models like GS-VIN, the increase in parameters leads to unstable network training despite the reduced network depth.
In summary, for tasks within smaller domains, we recommend using $f=5$ or $f=7$. 
However, when applying VIN-based models to larger domains, such as inputs of $64\times64$ or above, the required number of iterations for the network also increases. 
In this case, increasing the value of $f$ can be considered to reduce the network depth and thereby enhance the stability of network training.

Equation \eqref{eq:mfk} is a heuristic function that has been experimentally demonstrated to be effective to a certain extent and offers valuable insights for researchers working with VI-based models. 
By using \eqref{eq:mfk} as a foundation, researchers can fine-tune the $k,f$ parameters according to the domain size and model complexity of a specific task, ultimately enhancing model performance, reducing training time, and promoting training stability while effectively decreasing the trial-and-error cost.
However, it also highlights that there is still considerable room for improvement in \eqref{eq:mfk} as a heuristic function for adaptive iteration strategies. 
At this stage, researchers still need to manually adjust parameters according to varying circumstances. 
In future work, we will continue to refine and explore \eqref{eq:mfk} in order to identify more universal and effective heuristic approaches, thereby further enhancing the performance of our models in the AI domain.

\section{Conclusion}\label{sec: Conclusion}
In this paper, we propose a novel end-to-end planning model, GS-VIN. 
GS-VIN effectively performs accurate planning from image inputs.
GS-VIN offers two primary improvements over traditional VI-based models: 1) utilizing an adaptive iteration strategy to reduce planning iterations, decreasing error accumulation, and improving training stability;
and 2) introducing a gated summarization module, enabling better planning process summarization, further reducing single-iteration errors, and enhancing network performance.
We compare GS-VIN with three baseline methods (including VIN, VIRN, and DB-CNN) in terms of performance in traditional 2D grid-world path-finding tasks and the Mr. Pac-man game. 
Empirical results indicate that GS-VIN outperforms baseline methods in most cases, with the performance advantage increases as the size of the network input grows.
Additionally, we point out that the adaptive iteration strategy can be applied to the majority of VI-based models, and we provide a heuristic function with a significant reference value to assist researchers in optimizing VI-based models.
We demonstrate the effectiveness of this heuristic function through experiments on GS-VIN, VIN, and VIRN.

Despite the significant progress made by GS-VIN, the overall depth of the network is relatively deep due to the presence of the GS module, which can lead to more unstable training. 
Furthermore, we believe that there is room for improvement in the heuristic function of the adaptive iteration strategy, e.g., a heuristic function that takes into account the planning ability of the model itself.
In future work, we will continue to address these challenges and explore how to further enhance the model's performance while maintaining high training stability.



\bibliographystyle{IEEEtran}
\bibliography{ref}

\begin{thebibliography}{10}
\providecommand{\url}[1]{#1}
\csname url@samestyle\endcsname
\providecommand{\newblock}{\relax}
\providecommand{\bibinfo}[2]{#2}
\providecommand{\BIBentrySTDinterwordspacing}{\spaceskip=0pt\relax}
\providecommand{\BIBentryALTinterwordstretchfactor}{4}
\providecommand{\BIBentryALTinterwordspacing}{\spaceskip=\fontdimen2\font plus
\BIBentryALTinterwordstretchfactor\fontdimen3\font minus
  \fontdimen4\font\relax}
\providecommand{\BIBforeignlanguage}[2]{{%
\expandafter\ifx\csname l@#1\endcsname\relax
\typeout{** WARNING: IEEEtran.bst: No hyphenation pattern has been}%
\typeout{** loaded for the language `#1'. Using the pattern for}%
\typeout{** the default language instead.}%
\else
\language=\csname l@#1\endcsname
\fi
#2}}
\providecommand{\BIBdecl}{\relax}
\BIBdecl

\bibitem{dijkstra}
E.~W. Dijkstra, ``A note on two problems in connexion with graphs,'' in
  \emph{Edsger Wybe Dijkstra: His Life, Work, and Legacy}, 2022, pp. 287--290.

\bibitem{Astar}
P.~E. Hart, N.~J. Nilsson, and B.~Raphael, ``A formal basis for the heuristic
  determination of minimum cost paths,'' \emph{IEEE Transactions on Systems
  Science and Cybernetics}, vol.~4, no.~2, pp. 100--107, 1968.

\bibitem{VIN}
A.~Tamar, Y.~WU, G.~Thomas, S.~Levine, and P.~Abbeel, ``Value iteration
  networks,'' in \emph{Advances in Neural Information Processing Systems},
  D.~Lee, M.~Sugiyama \emph{et~al.}, Eds., vol.~29.\hskip 1em plus 0.5em minus
  0.4em\relax Curran Associates, Inc., 2016.

\bibitem{Bellman1957}
R.~Bellman, \emph{Dynamic Programming}.\hskip 1em plus 0.5em minus 0.4em\relax
  Princeton University Press, 1957.

\bibitem{LeCun1998}
Y.~LeCun, L.~Bottou, Y.~Bengio, and P.~Haffner, ``Gradient-based learning
  applied to document recognition,'' \emph{Proceedings of the IEEE}, vol.~86,
  no.~11, pp. 2278--2324, 1998.

\bibitem{DVIN}
X.~Jin, W.~Lan, T.~Wang, and P.~Yu, ``Value iteration networks with double
  estimator for planetary rover path planning,'' \emph{Sensors (Basel,
  Switzerland)}, vol.~21, 2021.

\bibitem{TVIN}
\BIBentryALTinterwordspacing
J.~Shen, H.~H. Zhuo, J.~Xu, B.~Zhong, and S.~J. Pan, ``Transfer value iteration
  networks,'' in \emph{The Thirty-Fourth {AAAI} Conference on Artificial
  Intelligence}.\hskip 1em plus 0.5em minus 0.4em\relax {AAAI} Press, 2020, pp.
  5676--5683. [Online]. Available:
  \url{https://ojs.aaai.org/index.php/AAAI/article/view/6022}
\BIBentrySTDinterwordspacing

\bibitem{AVIN}
D.~Schleich, T.~Klamt, and S.~Behnke, ``Value iteration networks on multiple
  levels of abstraction,'' \emph{Robotics: Science and Systems XV}, vol.
  abs/1905.11068, 2019.

\bibitem{UVIN}
L.~Zhang, X.~Li, S.~Chen, H.~Zang, J.~Huang, and M.~Wang, ``Universal value
  iteration networks: When spatially-invariant is not universal,'' in
  \emph{AAAI}, 2020.

\bibitem{Nardelli2018ValuePN}
N.~Nardelli, G.~Synnaeve, Z.~Lin, P.~Kohli, P.~H.~S. Torr, and N.~Usunier,
  ``Value propagation networks,'' \emph{ArXiv}, vol. abs/1805.11199, 2018.

\bibitem{Pflueger2018RoverIRLIR}
M.~Pflueger, A.~Agha, and G.~S. Sukhatme, ``Rover-irl: Inverse reinforcement
  learning with soft value iteration networks for planetary rover path
  planning,'' \emph{IEEE Robotics and Automation Letters}, vol.~4, pp.
  1387--1394, 2018.

\bibitem{sykora2020multi}
Q.~Sykora, M.~Ren, and R.~Urtasun, ``Multi-agent routing value iteration
  network,'' in \emph{International Conference on Machine Learning}.\hskip 1em
  plus 0.5em minus 0.4em\relax PMLR, 2020, pp. 9300--9310.

\bibitem{rumelhart1986learning}
D.~E. Rumelhart, G.~E. Hinton, and R.~J. Williams, ``Learning representations
  by back-propagating errors,'' \emph{nature}, vol. 323, no. 6088, pp.
  533--536, 1986.

\bibitem{hochreiter2001gradient}
S.~Hochreiter, Y.~Bengio, P.~Frasconi, J.~Schmidhuber \emph{et~al.}, ``Gradient
  flow in recurrent nets: the difficulty of learning long-term dependencies,''
  2001.

\bibitem{VIRN}
C.~Jinyu, L.~Jialong, M.~Zhenyu, and T.~Kenji, ``Value iteration residual
  network with self-attention,'' in \emph{22nd International Conference on
  Intelligent Systems Design and Applications}, 2022.

\bibitem{mnih2014recurrent}
V.~Mnih, N.~Heess, A.~Graves \emph{et~al.}, ``Recurrent models of visual
  attention,'' \emph{Advances in neural information processing systems},
  vol.~27, 2014.

\bibitem{Yang2018LearningUN}
S.~Yang, J.~Li, J.~Wang, Z.~Liu, and F.~Yang, ``Learning urban navigation via
  value iteration network,'' \emph{2018 IEEE Intelligent Vehicles Symposium
  (IV)}, pp. 800--805, 2018.

\bibitem{Li2021DynamicVI}
W.~Li, B.~Yang, G.~hua Song, and X.~Jiang, ``Dynamic value iteration networks
  for the planning of rapidly changing uav swarms,'' \emph{Frontiers of
  Information Technology \& Electronic Engineering}, vol.~22, pp. 687 -- 696,
  2021.

\bibitem{LSTM}
\BIBentryALTinterwordspacing
S.~Hochreiter and J.~Schmidhuber, ``{Long Short-Term Memory},'' \emph{Neural
  Computation}, vol.~9, no.~8, pp. 1735--1780, 11 1997. [Online]. Available:
  \url{https://doi.org/10.1162/neco.1997.9.8.1735}
\BIBentrySTDinterwordspacing

\bibitem{RNN}
D.~E. Rumelhart, G.~E. Hinton, and R.~J. Williams, ``Learning internal
  representations by error propagation,'' California Univ San Diego La Jolla
  Inst for Cognitive Science, Tech. Rep., 1985.

\bibitem{graves2006connectionist}
A.~Graves, S.~Fern{\'a}ndez, F.~Gomez, and J.~Schmidhuber, ``Connectionist
  temporal classification: labelling unsegmented sequence data with recurrent
  neural networks,'' in \emph{Proceedings of the 23rd international conference
  on Machine learning}, 2006, pp. 369--376.

\bibitem{amodei2016deep}
D.~Amodei, S.~Ananthanarayanan, R.~Anubhai, J.~Bai, E.~Battenberg, C.~Case,
  J.~Casper, B.~Catanzaro, Q.~Cheng, G.~Chen \emph{et~al.}, ``Deep speech 2:
  End-to-end speech recognition in english and mandarin,'' in
  \emph{International conference on machine learning}.\hskip 1em plus 0.5em
  minus 0.4em\relax PMLR, 2016, pp. 173--182.

\bibitem{sundermeyer2012lstm}
M.~Sundermeyer, R.~Schl{\"u}ter, and H.~Ney, ``Lstm neural networks for
  language modeling,'' in \emph{Thirteenth annual conference of the
  international speech communication association}, 2012.

\bibitem{radford2018improving}
A.~Radford, K.~Narasimhan, T.~Salimans, I.~Sutskever \emph{et~al.}, ``Improving
  language understanding by generative pre-training,'' 2018.

\bibitem{donahue2015long}
J.~Donahue, L.~Anne~Hendricks, S.~Guadarrama, M.~Rohrbach, S.~Venugopalan,
  K.~Saenko, and T.~Darrell, ``Long-term recurrent convolutional networks for
  visual recognition and description,'' in \emph{Proceedings of the IEEE
  conference on computer vision and pattern recognition}, 2015, pp. 2625--2634.

\bibitem{srivastava2015unsupervised}
N.~Srivastava, E.~Mansimov, and R.~Salakhudinov, ``Unsupervised learning of
  video representations using lstms,'' in \emph{International conference on
  machine learning}.\hskip 1em plus 0.5em minus 0.4em\relax PMLR, 2015, pp.
  843--852.

\bibitem{ConvLSTM}
X.~Shi, Z.~Chen, H.~Wang, D.-Y. Yeung, W.-K. Wong, and W.-c. Woo,
  ``Convolutional lstm network: A machine learning approach for precipitation
  nowcasting,'' \emph{Advances in neural information processing systems},
  vol.~28, 2015.

\bibitem{DB-CNN}
J.~Zhang, Y.~Xia, and G.~Shen, ``A novel learning-based global path planning
  algorithm for planetary rovers,'' \emph{Neurocomputing}, vol. 361, pp.
  69--76, 2018.

\bibitem{bengio1994learning}
Y.~Bengio, P.~Simard, and P.~Frasconi, ``Learning long-term dependencies with
  gradient descent is difficult,'' \emph{IEEE transactions on neural networks},
  vol.~5, no.~2, pp. 157--166, 1994.

\bibitem{Ding2022ScalingUY}
X.~Ding, X.~Zhang, Y.~Zhou, J.~Han, G.~Ding, and J.~Sun, ``Scaling up your
  kernels to 31×31: Revisiting large kernel design in cnns,'' \emph{2022
  IEEE/CVF Conference on Computer Vision and Pattern Recognition (CVPR)}, pp.
  11\,953--11\,965, 2022.

\bibitem{thrun1993issues}
S.~Thrun and A.~Schwartz, ``Issues in using function approximation for
  reinforcement learning,'' in \emph{Proceedings of the Fourth Connectionist
  Models Summer School}, vol. 255.\hskip 1em plus 0.5em minus 0.4em\relax
  Hillsdale, NJ, 1993, p. 263.

\bibitem{DDQN}
H.~V. Hasselt, A.~Guez, and D.~Silver, ``Deep reinforcement learning with
  double q-learning,'' \emph{ArXiv}, vol. abs/1509.06461, 2016.

\bibitem{DQN}
V.~Mnih, K.~Kavukcuoglu, D.~Silver \emph{et~al.}, ``Human-level control through
  deep reinforcement learning,'' \emph{Nature}, vol. 518, pp. 529--533, 2015.

\end{thebibliography}

\EOD

\end{document}